\documentclass[letterpaper, 10 pt, conference]{ieeeconf}  

\IEEEoverridecommandlockouts                              

\usepackage[utf8]{inputenc}
\usepackage[pdftex]{graphicx}
\usepackage[hidelinks]{hyperref}
\usepackage[table,xcdraw]{xcolor}
\title{
Challenges in Visual Anomaly Detection for Mobile Robots
}
\author{Dario Mantegazza, Alessandro Giusti, Luca M. Gambardella, Andrea Rizzoli and J\'er\^ome Guzzi 
\thanks{Dario Mantegazza, Alessandro Giusti, Luca M. Gambardella, Andrea Rizzoli and J\'er\^ome Guzzi are with the Dalle Molle Institute for Artificial Intelligence (IDSIA), USI-SUPSI, Lugano, Switzerland.
This work was supported as a part of NCCR Robotics, a National Centre of Competence in Research, funded by the Swiss National Science Foundation (grant number 51NF40\_185543) and by the European Commission through the Horizon 2020 project 1-SWARM, grant ID 871743.}
}

\date{March 2022}

\begin{document}

\maketitle
\begin{abstract}
We consider the task of detecting anomalies for autonomous mobile robots based on vision.
We categorize relevant types of visual anomalies and discuss how they can be detected by unsupervised deep learning methods.
We propose a novel dataset built specifically for this task, on which we test a state-of-the-art approach; we finally discuss deployment in a real scenario.  

\end{abstract}
\section{Introduction}


Autonomous robots are sometimes deployed in environments that include hazards, i.e., locations that might disrupt the robot's operation, possibly causing it to crash, get stuck, and more generally to fail its mission.
Robots are usually capable to perceive hazards that are expected during system development, and that can be explicitly accounted for when designing the perception subsystem. Nonetheless, during deployment a robot might incur in situations that were not expected during system design (anomalies). 

In this paper, we discuss the challenges related to detecting anomalies from visual inputs and provide a dataset including many anomaly types.  Because we don't have any model of how these hazards might appear, we consider anything that is novel or unusual as a potential hazard to be avoided.
We do not deal with the problem of choosing an appropriate reaction once an anomaly is detected, which depends on the specific scenario.


Ruff~\cite{ruff2021unifying} states that an anomaly is ``an observation that deviates considerably from some concept of normality''.  This definition highlights the importance of the context: an anomaly is defined only when a concept of normality exists.  In the literature, anomalies have taken different meanings depending on the applications: unusual patterns in flight data~\cite{birnbaum2015unmanned}, texture changes in manufactured products~\cite{haselmann2018anomaly}, unexpected obstacles in cultivated fields~\cite{christiansen2016deepanomaly}. Surveys~\cite{chandola2009anomaly, ruff2021unifying} on Anomaly Detection propose a general categorization of anomalies, based on two criteria: one classifying point, contextual and collective anomalies;
the other distinguishing between low-level sensory and high-level semantic anomalies.

\begin{figure*}[!ht]
    \centering
    \includegraphics[width=\textwidth]{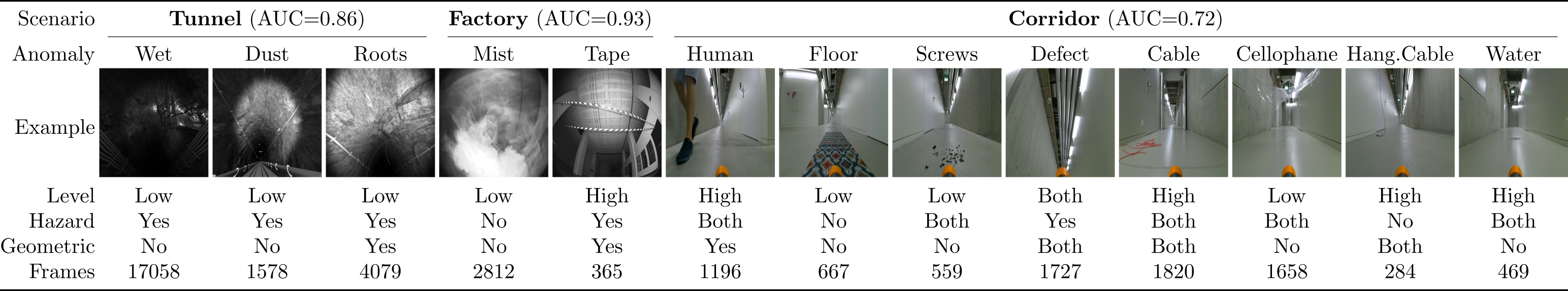}
    \caption{Summary of the anomalies represented in the dataset, with an example of each. Top row reports the score of our model for each scenario. Below: the categorization of the dataset's anomalies along three of the four proposed axes.}
    \label{fig:tabellozza}
\end{figure*}
\subsection{Categorization of anomalies}
We propose to categorize visual anomalies encountered by mobile robots during their operation along four independent axes.

The first, following Ruff et~al.~\cite{ruff2021unifying}, differentiates low-level sensory and high-level semantic anomalies.  Low-level anomalies are described by features close to the image space, such as image brightness, smoothness, noise, and texture; one example of such anomaly occurs when the robot suddenly finds itself in the dark or when it is blinded by direct light.  High-level anomalies refer to the semantic contents of the image: examples include the observation of a pressure gauge reporting a different value than usual, or of a puddle of liquid on the ground.

The second axis represents whether an anomaly is a hazard to the robot.  This is specific to the robot's characteristics (an oil puddle on the floor might represent a hazard for a ground robot but not for a drone) and its task (the puddle is not dangerous unless it lies on the robot's path).

The third axis differentiates anomalies that are relevant to the robot mission and anomalies that are not. For example, a patrolling robot might want to detect and report the fact that a door that is usually closed is observed to be open, whereas a delivery robot should not be affected by this observation unless it impacts its path planning.

The fourth axis discriminates visual anomalies that are geometric in nature from those that are not. Geometric anomalies have a well-defined 3D shape in the robot environment (e.g. a never-before-seen object in a normally-free corridor) or consist of changes in the position or shape of a part of the environment (a wall that collapsed).  Non-geometric anomalies are anomalies that, while perceivable using an RGB camera, would be undetectable with an ideal depth sensor: for example, a puddle on the ground; plaster rubble scattered across a building floor; dust, fog, or smoke; a wet ceiling in a tunnel.

\section{A Visual Anomaly Dataset for Robotics}

Any machine learning model requires a large, representative dataset to be trained and evaluated on.
Since dataset collection is expensive, visual anomaly detection literature often relies on existing classification datasets such as MNIST~\cite{lecun1998gradient}, ImageNet~\cite{deng2009imagenet}, CIFAR~\cite{krizhevsky2009learning};
a set of classes is selected as normal while the others, often with synthetic variations, represent anomalies.
This approach has major limitations because it does not capture the visual characteristics of realistic anomalies with respect to normal data.  This is solved with task-specific datasets, which have been proposed for industrial inspection and healthcare.  We introduce a new dataset~\footnote{The dataset is available here \url{https://github.com/idsia-robotics/hazard-detection}} that is specific to visual anomaly detection for mobile robots.
In the robotic field, to the best of our knowledge, there is no dataset publicly available for anomaly detection.


Our dataset covers three different scenarios.  Data for each scenario is divided into three sets: a training and a validation set, composed exclusively of normal samples; and a testing set, composed of normal and anomalous samples;  each anomalous sample is annotated with the exact type of anomaly that it represents. 

The scenarios are: \emph{Tunnel}, a simulation of a drone flying inside an underground tunnel, with 3 kinds of anomalies; \emph{Factory}, recordings from a real drone inside a factory, with 2 types of anomalies;  \emph{Corridor}, a real wheeled mobile robot moving inside some university's corridors, with 8 anomaly types.
All the datasets are recorded from the front-facing camera of the robots at 30 frames per second. For the experiments, all samples are reshaped to square images with size $64\times64$ pixels.
In Figure~\ref{fig:tabellozza} we show examples of all the represented anomalies. We also classify anomaly types along the axes introduced before. The third axis is not represented in the Figure since it is mission dependent and our dataset is gathered with no specific task in mind.

\section{Experiments and Perspectives}

Anomaly detection can be defined as a binary classification problem, where each sample is associated with one of two classes: normal samples are classified as negative and anomalies as positive.

The problem could be solved using a supervised machine learning approach that relies on a labeled train set that contains examples of both classes.
However, in our definition anomalies are \emph{rare} and \emph{unexpected} events.  Because of this, collecting a large training set of anomalies would be very time-consuming, and collecting a representative training set of all possible anomalies is unfeasible.

Therefore, we focus our analysis on \emph{unsupervised} methods.  
In this setting, the anomaly detector is learned from a training set composed exclusively of normal samples.  When used for inference, the anomaly detection model will return  an anomaly score for each sample that is low for normal samples and high for hazardous ones. State-of-the-art visual anomaly detection models rely on Deep Learning techniques to learn a similarity metric that accounts for the expected variability in the normal training images.
We developed an anomaly detector based on Autoencoder~\cite{kramer1992autoassociative} and Real NVP~\cite{dinh2016density}, that takes as input an image and produces an anomaly score.
The model is trained only on normal samples and is tested on both normal and anomalous images. 

From anomaly scores computed on a testing set with normal and anomalous samples, we measure the quality of the detector using the Area Under the ROC Curve (AUC) as a metric.  The AUC value ranges from 0 to 1 where 1 is a perfect anomaly detector and 0.5 equals a random classifier.
%
On the top row of Figure~\ref{fig:tabellozza} we show the AUC obtained by our model tested on the three scenarios.

\begin{figure}[!htp]
    \centering
    \includegraphics[width=0.8\columnwidth]{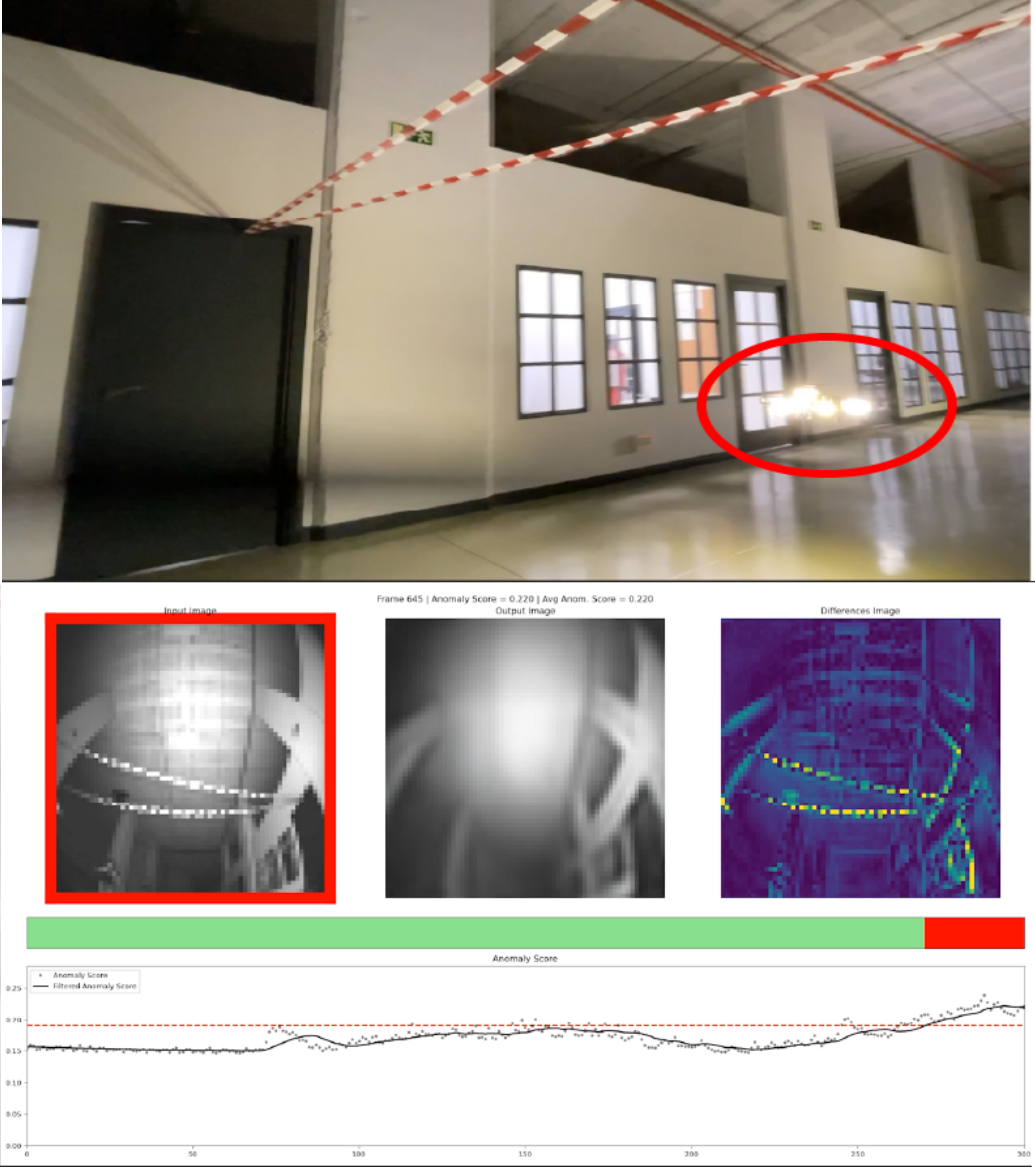}
    \caption{Deployment in the Factory scenario: the drone advances normally until it detects an anomaly (a tape crossing its path); a time series of anomaly scores in previous frames is reported at the bottom (see supplementary video \url{https://youtu.be/SylhxUl20C0}).}
    \label{fig:drone}
\end{figure}

We finally deployed the model trained on the \emph{Factory} dataset on an autonomous drone, that stops its mission and backtracks in case an anomaly is detected. Figure~\ref{fig:drone} illustrates that anomaly detection avoids collision with an unforeseen obstacle (a thin tape that would not be otherwise seen by the drone's obstacle detection suite).


 

We are currently focusing on two related topics: allowing pre-trained models to rapidly adapt to new environments via active learning and domain adaptation, and exploiting recordings of known anomalies 
through outlier exposure~\cite{hendrycks2018deep}.

\newpage
\bibliographystyle{IEEEtran}
\bibliography{biblio}
\end{document}